\documentclass[conference, letterpaper, 10pt]{ieeeconf}
\usepackage{ieee}
\usepackage{microtype}

\IEEEoverridecommandlockouts
\newcommand{\browserTab}[1]{
    \begin{tikzpicture}

    \fill[gray!80, rounded corners=0.3cm] (0, 0) rectangle (7, 0.6); %
    \draw[black!20, rounded corners=0.3cm] (0, 0) rectangle (7, 0.6); %

    \fill[red] (0.5, 0.3) circle (0.12); %
    \fill[yellow] (0.85, 0.3) circle (0.12); %
    \fill[green] (1.2, 0.3) circle (0.12); %

    \fill[white, rounded corners=0.3cm] (1.5, 0.1) rectangle (6.5, 0.5); %
    \draw[black!40, rounded corners=0.3cm] (1.5, 0.1) rectangle (6.5, 0.5); %

    \node[font=\footnotesize] at (4, 0.3) {#1}; %

    \end{tikzpicture}
}

\title{\LARGE \bf
ROS2WASM: Bringing the Robot Operating System to the Web}

\author{Tobias Fischer$^1$, Isabel Paredes$^{2,3}$, Michael Batchelor$^1$, Thorsten Beier$^2$, Jesse Haviland$^1$,\\Silvio Traversaro$^4$, Wolf Vollprecht$^5$, Markus Schmitz$^3$, and Michael Milford$^1$
\thanks{$^1$TF, MB, JH and MM are with the QUT Centre for Robotics, School of Electrical Engineering and Robotics, Queensland University of Technology, Brisbane, QLD 4000, Australia (e-mail: \{tobias.fischer, mj.batchelor, j.havl, michael.milford\}@qut.edu.au).}%
\thanks{$^2$IP and TB are with QuantStack, 16 Avenue Curti, 94100 Saint-Maur-des-Fossés, France (e-mail: \{isabel.paredes, thorsten.beier\}@quantstack.net).}%
\thanks{$^3$IP and MS are with the Institute of Mechanism Theory, Machine Dynamics and Robotics, RWTH Aachen, Germany (e-mail: schmitzm@igmr.rwth-aachen.de).}%
\thanks{$^4$ST is with the Fondazione Istituto Italiano Di Tecnologia (Italian Institute of Technology), Genova, Italy (e-mail: silvio@traversaro.it).}%
\thanks{$^5$WF is with prefix.dev, Französische Straße 12, 10117 Berlin, Germany (e-mail: w.vollprecht@gmail.com).}%
\thanks{This research was partially supported by funding from ARC DECRA Fellowship DE240100149 to TF, ARC Laureate Fellowship FL210100156 to MM, and the QUT Centre for Robotics.}%
}

\begin{document}
\maketitle
\thispagestyle{empty}
\pagestyle{empty}

\begin{abstract}
The Robot Operating System (ROS) has become the de facto standard middleware in robotics, widely adopted across domains ranging from education to industrial applications. The RoboStack distribution, a conda-based packaging system for ROS, has extended ROS's accessibility by facilitating installation across all major operating systems and architectures, integrating seamlessly with scientific tools such as PyTorch and Open3D. This paper presents ROS2WASM, a novel integration of RoboStack with WebAssembly, enabling the execution of ROS 2 and its associated software directly within web browsers, without requiring local installations. ROS2WASM significantly enhances the reproducibility and shareability of research, lowers barriers to robotics education, and leverages WebAssembly's robust security framework to protect against malicious code. We detail our methodology for cross-compiling ROS 2 packages into WebAssembly, the development of a specialized middleware for ROS 2 communication within browsers, and the implementation of \emph{www.ros2wasm.dev}, a web platform enabling users to interact with ROS 2 environments. Additionally, we extend support to the Robotics Toolbox for Python and adapt its Swift simulator for browser compatibility. Our work paves the way for unprecedented accessibility in robotics, offering scalable, secure, and reproducible environments that have the potential to transform educational and research paradigms.
\end{abstract}

\section{Introduction}
The Robot Operating System (ROS) has emerged as the de facto standard robotics middleware, facilitating communication between multiple nodes in a network~\cite{macenski2022robot}. Its extensive libraries and active community support have led to widespread adoption across domains ranging from education to industrial applications~\cite{estefo2019robot}. However, as the ROS user base expands, issues related to the accessibility of ROS distributions have surfaced. ROS2WASM addresses these concerns by bridging the gap between high-performance scientific computing and web-based applications, making advanced robotics tools more accessible to a broader audience. To aid readers in navigating the technical terms and concepts used throughout this paper, a glossary is provided in Table~\ref{tab:glossary}.

\begin{figure}[t]
    \centering
    \browserTab{ROS 2 @ localhost}
    \includegraphics[width=0.96\columnwidth]{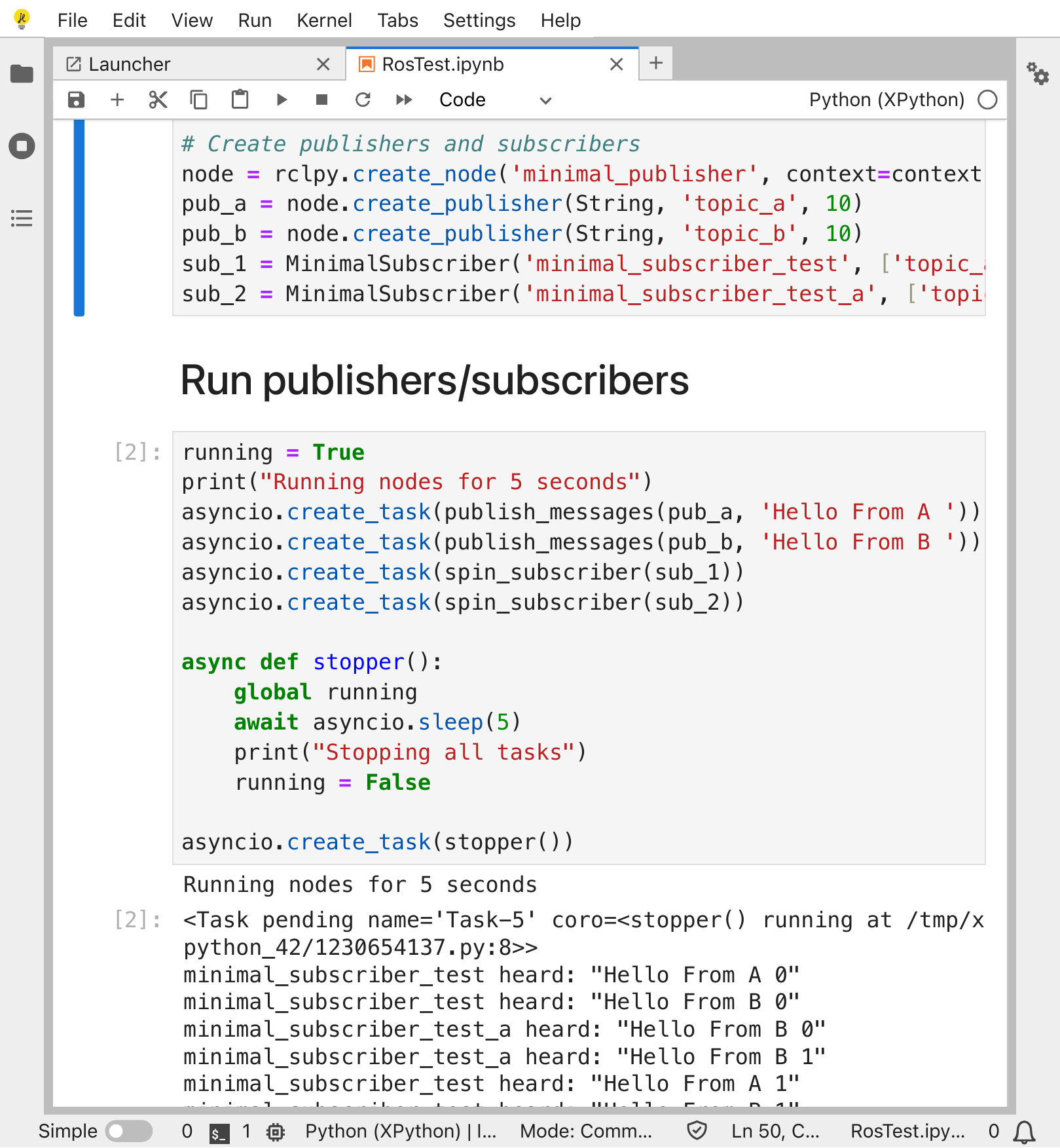}
    \caption{Execution of a ROS 2 publisher and subscriber setup within a Jupyter Notebook environment running entirely in the browser, with no local installation required (\url{www.ros2wasm.dev}). The notebook creates publishers for topics \texttt{topic\_a} and \texttt{topic\_b}, as well as two minimal subscribers listening to these topics. The log output at the bottom shows the concurrent execution of these nodes, with messages being published and received, verifying seamless communication between ROS 2 nodes directly within the browser-based interface.}
    \label{fig:jupyterlite}
    \vspace*{-0.25cm}
\end{figure}

ROS 2 has made significant strides by enabling native installations on major platforms, including Ubuntu, Windows, and macOS. The RoboStack effort has further extended ROS 2’s reach by allowing installations within \codefont{conda} environments~\cite{fischer2021robostack}. Nevertheless, variations in package versions across platforms can complicate the replication of workspaces, posing challenges for both educational and research applications~\cite{cervera2018try}.

Replicating workspaces is particularly beneficial in scenarios such as teaching robotics courses~\cite{kohut2023teaching} and reproducing research papers~\cite{cervera2023run}. Instructors often struggle to provide students with the necessary tools, resorting to maintaining a computer lab or providing instructions for workplace replication. However, students’ diverse computing environments can lead to inconsistencies in how ROS packages function due to differences in operating systems, library versions, and system configurations. ROS2WASM mitigates these issues by providing a standardized execution environment within the browser itself. By compiling ROS 2 to WebAssembly, students can interact with identical software configurations regardless of their hardware or operating system, running ROS 2 directly in their browsers (e.g., Chrome, Firefox, Safari, Edge). ROS2WASM thus eliminates any local setup or reliance on university servers (Figure~\ref{fig:jupyterlite}).

In research, reproducibility faces similar obstacles. Traditional environments are often rigid and difficult to adapt, and version compatibility issues can impede reproducibility efforts. While Docker containers provide a potential solution, installing Docker and managing containers can be daunting and time-consuming for many aspiring roboticists~\cite{cervera2019roslab,white2017ros}.

We propose a solution based on WebAssembly (WASM)~\cite{haas2017bringing,wang2021empowering}, a low-level bytecode format designed to run efficiently in web browsers. WASM is fast, secure, portable, and well-sandboxed across all major browsers, making it ideal for high-performance web applications.

Instead of relying on Docker images, we package \mbox{ROS 2} workspaces for web browsers, enabling easy distribution via simple links. These browser-based environments can run on any operating system and processor architecture, from smartphones to high-end desktops. By cross-compiling ROS packages to WebAssembly, ROS2WASM runs ROS nodes directly in the browser and facilitates communication between them. Unlike other ROS tools that depend on an existing ROS system or server connection, ROS2WASM creates a fully isolated, sandboxed environment within the browser. 

Although ROS2WASM involves initial complexity in cross-compilation and middleware development, this complexity is largely encapsulated on the developer side. For end users, ROS2WASM offers a lower barrier to entry compared to Docker and similar solutions. ROS2WASM builds upon RoboStack~\cite{fischer2021robostack}, which integrates ROS with the Conda package manager. RoboStack offers several key advantages, including reproducible environments, robust version management, compatibility across platforms and Linux distributions, distribution of pre-built binaries, environment isolation, and simplified package creation.

Our contributions are as follows:
\begin{enumerate}
    \item A custom ROS 2 middleware implementation, \codefont{rmw-wasm}, cross-compiled into WebAssembly modules, enabling communication between ROS 2 modules within browser environments. This middleware also enables communication with external physical robots running standard ROS 2 installations using \url{https://wiki.ros.org/roslibjs}.
    \item Cross-compilation of numerous ROS 2 packages to WebAssembly, including \codefont{rclpy}, to provide Python support for beginners.
    \item Deployment of \url{www.ros2wasm.dev}, which showcases an example browser environment for ROS2WASM.
    \item Cross-compilation of the Robotics Toolbox for Python and adaptation of its Swift simulator for compatibility with WebAssembly, providing a concrete example of adaptation that can serve as a foundation for further extensions by other users.
\end{enumerate}

  \begin{table*}[t]
    \renewcommand*{\arraystretch}{1.2}
    \centering
    \caption{Glossary}
    \vspace*{-0.1cm}
    \label{tab:glossary}
    \begin{tabularx}{1.0\linewidth}{
      | >{\hsize=.35\hsize\linewidth=\hsize}X
      >{\hsize=1.65\hsize\linewidth=\hsize}X |
    }
      \hline
      \textbf{Term} & \textbf{Description} \\
      \hline
         Boa & Build tool for conda packages. \\
         Conda & An open-source package management system and environment management system that allows users to install and manage packages, dependencies, and environments across different platforms.
         \\ Conda-forge & A community-driven platform providing a vast collection of conda packages, built and maintained using continuous integration (CI) services to ensure cross-platform compatibility.\\
         Cross-compiler & A compiler that generates executable code for a platform different from the one on which the compiler is running.\\
         DDS & Data Distribution Service, a middleware protocol and API standard for data exchange between nodes in a distributed system, ensuring reliable, real-time communication.
         \\ Emscripten & Compiler toolchain to convert C/C++ code into WebAssembly.\\
         Emscripten-forge & A repository of conda packages built using Emscripten, enabling the cross-compilation of C/C++ code to WebAssembly.\\
         \codefont{emscripten::val} & A utility provided by the Emscripten toolchain that facilitates the interaction between C++ and JavaScript by allowing C++ code to call and manipulate JavaScript objects and functions seamlessly within a WebAssembly environment.\\
         JupyterLite & A JupyterLab distribution which runs on a web browser.\\
         RoboStack & A project that integrates the Robot Operating System (ROS) with the conda package manager, enabling easy installation and management of ROS distributions across different platforms.\\
         Toolchain & A set of programming tools used together in a development process, typically including a compiler, linker, and debugger, among others, to automate the building and deployment of software.\\
         WebAssembly (WASM) & Low-level binary format designed for efficient execution on modern web browsers.\\
         Web Worker & A JavaScript API that allows running scripts in background threads, enabling concurrent execution without affecting the performance of the main browser thread.\\
      \hline
    \end{tabularx}
    \vspace*{-0.25cm}
  \end{table*}

\section{Related Works}
This section overviews existing technologies related to WebAssembly, ROS in the browser, and browser-based robotics environments that contextualize our work.

\subsection{Overview of ROS 2 and RoboStack}
Since its inception, ROS has been pivotal in advancing robotics research and development~\cite{quigley2009ros}. It has enabled the creation of complex robotic systems that can operate in diverse environments, from autonomous vehicles~\cite{hellmund2016robot} to industrial automation~\cite{he2022ros2}. The modular architecture of ROS, combined with its extensive library of packages, has made it a powerful tool for both prototyping and deploying robotic systems.

ROS 2~\cite{macenski2022robot}, the successor of ROS, builds on the strengths of its predecessor while addressing several limitations. It introduces a new communication middleware based on the Data Distribution Service (DDS), which provides real-time communication capabilities, improved security, and better support for multi-robot systems. DDS enables ROS 2 nodes to communicate over a network in a decentralized and scalable manner, making it ideal for distributed robotic applications.%

RoboStack~\cite{fischer2021robostack} is a project that integrates ROS with the Conda package manager, significantly enhancing the accessibility and usability of ROS across different platforms. By leveraging Conda, RoboStack allows users to install ROS distributions in isolated environments, ensuring reproducibility and compatibility across different operating systems and architectures. This approach simplifies the installation process, reduces dependency conflicts, and allows for easier management of ROS packages. RoboStack also facilitates the distribution of pre-built binaries, which can be cross-compiled for various platforms. This capability is central to the development of our proposed ROS2WASM, where we cross-compile ROS to WebAssembly to enable seamless deployment of ROS packages in a web environment.

\subsection{WebAssembly}
WebAssembly (WASM) is a relatively new technology in the web development landscape~\cite{haas2017bringing}, which emerged in 2017 based on a collaboration between major browser vendors -- Chrome, Edge, Firefox, and Safari.

WASM is a binary code format designed for a stack-based virtual machine, optimized for performance, security, efficiency, and portability. It can be executed within web browsers as a complement to JavaScript or independently in other environments. These characteristics make WASM ideal for high-performance applications, including games, scientific visualizations, simulations, and developer tools. Due to its capabilities, WASM is particularly well-suited for implementing a robotics environment in the browser.

Early demonstrations of WASM's potential included porting the game AngryBots, developed by Unity Technologies, to the web~\cite{angrybots}. Similarly, Doom 3 was ported to WASM using Emscripten~\cite{d3wasm}, showcasing the ability to run large, demanding C++ programs within a browser. WASM's adaptability and performance make it appealing for robotics applications.

\begin{figure}[t]
    \centering
    \includegraphics[width=0.99\columnwidth]{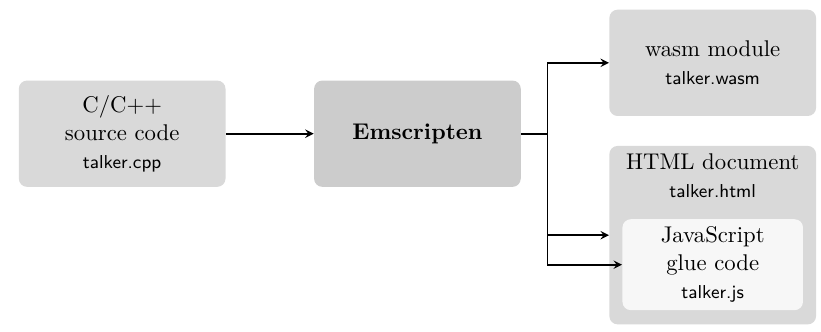}
    \caption{Overview of the pipeline for compiling C/C++ source code into WebAssembly (WASM) using Emscripten. The process begins with the C/C++ source code (e.g., \texttt{talker.cpp}), which is passed through the Emscripten compiler. Emscripten converts this source code into three primary outputs: a WebAssembly module (\texttt{talker.wasm}), an HTML document (\texttt{talker.html}) that serves as the interface to load and run the WASM module, and JavaScript glue code (\texttt{talker.js}) that manages the interaction between the WebAssembly module and the web environment. This pipeline allows native C/C++ applications to be executed in the browser, offering near-native performance and seamless integration with web technologies.}
    \label{fig:c_to_wasm}
    \vspace*{-0.2cm}
\end{figure}

For ROS2WASM, we use Emscripten~\cite{zakai2011emscripten}, an open-source compiler toolchain for converting C/C++ code to WebAssembly. Emscripten supports the compilation of various languages into WASM by leveraging LLVM~\cite{LLVM}, enabling their execution on the Web, Node.js, or other WASM runtimes. The pipeline for compiling C/C++ code to WASM with Emscripten is illustrated in Figure~\ref{fig:c_to_wasm}. Additionally, Emscripten can compile C/C++ runtimes of other languages such as Python, facilitating the indirect execution of those languages in the browser. Specifically, we build on top of emscripten-forge, a project that implements a build pipeline to compile Conda packages to Web Assembly. Emscripten-forge makes use of the same build tools as RoboStack, making it well suited to cross-compile ROS 2 to Web Assembly.

\subsection{ROS in the Browser}
The closest analogue to ROS2WASM is the ROS on Web project~\cite{rosonweb}, which shares similar goals of enabling ROS nodes to run entirely within a browser by cross-compiling C++ code to WebAssembly and using web workers for internal communication. However, a significant limitation of ROS on Web is its closed-source nature, which restricts access to the implementation details. While the project indicates that it replaces the ROS 2 middleware with a custom design and employs web workers, the exact methods remain undisclosed.

Another relevant project is the roswasm suite~\cite{roswasm_suite}, an open-source collection of libraries designed to assist in cross-compiling C++ ROS nodes to WebAssembly. However, the roswasm suite is primarily focused on the outdated ROS 1 distribution, limiting its applicability to ROS 2 users.

\subsection{Browser-Based Robotics Environments}
Several other projects, while not directly aligned with ROS2WASM's objectives, are relevant to the broader context of developing robotics environments for web browsers. The Robot Web Tools website (\url{https://robotwebtools.github.io}) offers a comprehensive overview of open-source libraries and tools that facilitate the integration of robotic frameworks with the web. ROSWeb~\cite{rosweb} aggregates all the ROS widgets provided by Robot Web Tools into a single web application, offering an interactive and user-friendly interface that requires minimal programming expertise.

One of the leading examples in this domain is Foxglove Studio~\cite{foxglove}, which focuses on the visualization and debugging of robotic tasks. Although Foxglove Studio's primary focus is observability rather than direct robotic control, its cross-platform support for ROS 1 and ROS 2 and its availability as both a desktop application and a browser-based tool make it a significant contribution to the field. However, Foxglove Studio functions as an add-on to existing ROS installations rather than a standalone solution. Furthermore, Foxglove Studio is no longer open source and free for general use.

The rosbridge suite~\cite{crick2017rosbridge} enables non-ROS systems to communicate with ROS via JSON, supporting both \mbox{ROS 1} and ROS 2 but requiring existing ROS installations, thus not being a complete alternative for running ROS in the browser. The ROS Control Center~\cite{ros_control_center} is a web application that uses rosbridge to establish a WebSocket connection to a robotics system running ROS. In contrast, ROSboard~\cite{rosboard} does not depend on the rosbridge suite. Instead, it implements a custom Tornado web server that functions as both a web server and a WebSocket server, providing a lightweight and streamlined interface for interacting with ROS systems.

\section{Methods and Demonstrations}
This section details the methodology employed to develop ROS2WASM. We begin with the design of a custom middleware architecture for ROS 2 with WebAssembly, enabling communication within browser environments (Section~\ref{sec:middleware}). Next, we describe the cross-compilation process extending RoboStack to support WebAssembly (Section~\ref{sec:crosscompile}), followed by the web integration strategy using JupyterLite for browser-based access (Section~\ref{sec:webintegration}). Additionally, we cover adaptations to the Robotics Toolbox for Python and Swift Robotics Simulator for the sandboxed web environment (Section~\ref{sec:swift}). Finally, we discuss methods for communicating with non-ROS physical robots, demonstrating a practical application of ROS2WASM (Section~\ref{sec:vernie}).

\subsection{Custom Middleware Architecture}
\label{sec:middleware}

\begin{figure}[t]
    \centering
    \includegraphics[width=0.99\columnwidth]{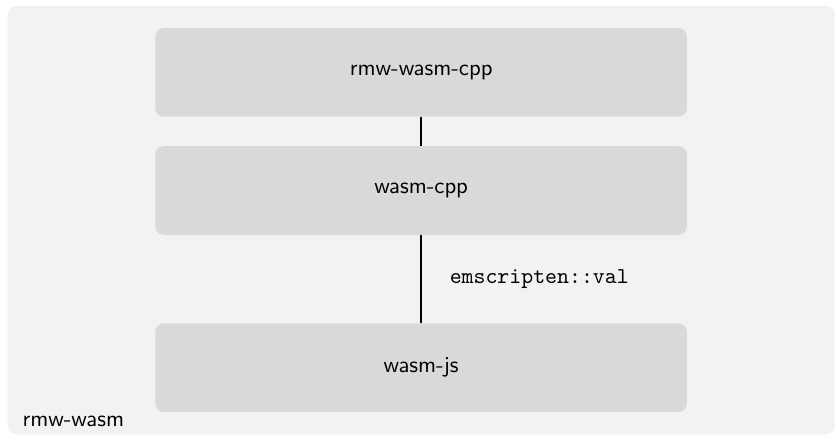}
    \vspace*{-0.2cm}
    \caption{Architecture of the \texttt{rmw-wasm} middleware enabling ROS 2 communication in the browser. The system comprises three main components: \texttt{rmw-wasm-cpp} (interfaces ROS 2 packages with the WebAssembly-based middleware), \texttt{wasm-cpp} (the C++ core that processes messages and communicates with JavaScript), and \texttt{wasm-js} (manages network participants and message traffic within the browser). Communication between \texttt{wasm-cpp} and \texttt{wasm-js} is facilitated by \texttt{emscripten::val}, allowing seamless integration between C++ and JavaScript.}
    \vspace*{-0.2cm}
    \label{fig:rmw-wasm}
\end{figure}

Our custom middleware architecture comprises three main packages: \codefont{rmw-wasm-cpp}, \codefont{wasm-cpp}, and \codefont{wasm-js} (Figure~\ref{fig:rmw-wasm}). This design draws inspiration from \codefont{rmw-email}~\cite{ros_over_email}, which contains a middleware for sending and receiving strings over email, and an RMW implementation that allows ROS 2 to use this middleware for message exchange.

The \codefont{rmw-wasm-cpp} package acts as an adapter between ROS 2 packages and the middleware implementation; it contains all the functions required by the \codefont{rmw} interface. A primary function of this package is to convert binary ROS messages into YAML strings using dynamic message introspection (\codefont{dynmsg}). It then passes these strings to the \codefont{wasm-cpp} package for further processing. Conversely, when messages are retrieved by subscribers, \codefont{rmw-wasm-cpp} converts the YAML string messages back to binary ROS messages to return to the requester.

The middleware implementation is divided into \codefont{wasm-cpp} (implemented in C++) and \codefont{wasm-js} (implemented in JavaScript). Communication between these two packages is facilitated by \codefont{emscripten::val}, which allows JavaScript functions to be called from C++ and vice versa~\cite{zakai2011emscripten}.

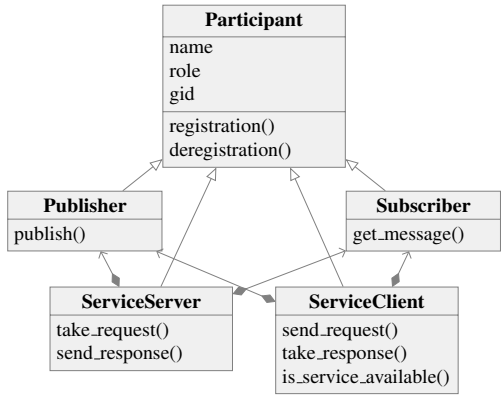
\begin{figure}[t]
    \centering
    \scalebox{0.75}{
    \begin{tikzpicture}%
        \begin{class}[text width=3cm]{Participant}{0,0}
            \attribute{name}
            \attribute{role}
            \attribute{gid }
            \operation{registration()}
            \operation{deregistration()}
        \end{class}

        \begin{class}[text width=2.5cm]{Publisher}{-3,-3.3}
            \inherit{Participant}
            \operation{publish()}
        \end{class}

        \begin{class}[text width=2.5cm]{Subscriber}{3,-3.3}
            \inherit{Participant}
            \operation{get\_message()}
        \end{class}

        \begin{class}[text width=3cm]{ServiceServer}{-2,-5}
            \inherit{Participant}
            \operation{take\_request()}
            \operation{send\_response()}
        \end{class}

        \composition{ServiceServer}{}{}{Publisher}
        \composition{ServiceServer}{}{}{Subscriber}

        \begin{class}[text width=3cm]{ServiceClient}{2,-5}
            \inherit{Participant}
            \operation{send\_request()}
            \operation{take\_response()}
            \operation{is\_service\_available()}
        \end{class}

        \composition{ServiceClient}{}{}{Publisher}
        \composition{ServiceClient}{}{}{Subscriber}

    \end{tikzpicture}
    }
    \caption{Class diagram of the \texttt{wasm-cpp} package, showing the relationships between ROS 2 entities. The Participant class is the base, with \texttt{Publisher}, \texttt{Subscriber}, \texttt{ServiceServer}, and \texttt{ServiceClient} as derived classes. Each class handles specific ROS 2 functions, such as publishing, subscribing, and managing service requests and responses, enabling communication within the WebAssembly environment.}
    \label{fig:classes}
\end{figure}

The \codefont{wasm-cpp} package serves as the core of the implementation and functions as a bridge to JavaScript modules (Figure~\ref{fig:classes}). In this package, ROS elements are constructed hierarchically, with the smallest unit being a participant. Any ROS subscriber, publisher, service server, service client, action server, or action client is modeled as a participant with a specific role. Upon initialization, each participant receives a unique identifier for tracking purposes.

\begin{figure}[t]
    \centering
    \scalebox{0.8}{
    \begin{tikzpicture}
        \node (clt) [
            rectangle,
            rounded corners,
            minimum width=2.5cm,
            text depth=3cm,
            xshift=-3cm,
            fill=gray!20,
        ] {Service Client};

        \node (cltpub) [
            rectangle,
            rounded corners,
            minimum width=2cm,
            minimum height=1cm,
            xshift=-3cm,
            yshift=0.5cm,
            fill=gray!5,
        ] {Publisher};

        \node (cltsub) [
            rectangle,
            rounded corners,
            minimum width=2cm,
            minimum height=1cm,
            xshift=-3cm,
            yshift=-1cm,
            fill=gray!5,
        ] {Subscriber};

        \node (srv) [
            rectangle,
            rounded corners,
            minimum width=2.5cm,
            text depth=3cm,
            xshift=3cm,
            fill=gray!20,
        ] {Service Server};

        \node (srvsub) [
            rectangle,
            rounded corners,
            minimum width=2cm,
            minimum height=1cm,
            xshift=3cm,
            yshift=0.5cm,
            fill=gray!5,
        ] {Subscriber};

        \node (srvpub) [
            rectangle,
            rounded corners,
            minimum width=2cm,
            minimum height=1cm,
            xshift=3cm,
            yshift=-1cm,
            fill=gray!5,
        ] {Publisher};

        \node (t1) [
            rectangle,
            rounded corners,
            minimum width=2cm,
            minimum height=1cm,
            yshift=0.5cm,
        ] {\codefont{/request\underscore topic}};

        \node (t2) [
            rectangle,
            rounded corners,
            minimum width=2cm,
            minimum height=1cm,
            yshift=-1cm,
        ] {\codefont{/response\underscore topic}};

        \draw [arrow] (cltpub) -- (t1);
        \draw [arrow] (t1) -- (srvsub);
        \draw [arrow] (srvpub) -- (t2);
        \draw [arrow] (t2) -- (cltsub);

    \end{tikzpicture}
    }
    \caption{Data flow between a ROS service client and server within \texttt{wasm-cpp}. The service client initiates a request by publishing to the \texttt{/request\_topic}, which is received by the service server's subscriber. The server processes the request and then publishes a response to the \texttt{/response\_topic}, which is received by the client's subscriber. This setup demonstrates how service clients and servers in ROS are built on top of publisher-subscriber pairs, ensuring communication via specific request and response topics.}
    \vspace*{-0.2cm}
    \label{fig:service}
\end{figure}
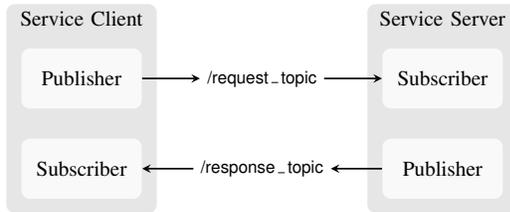

Publishers and subscribers are the simplest participant types and are responsible for sending and receiving messages, respectively. Service participants are derived from these basic types; service clients and servers consist of a publisher-subscriber pair (Figure~\ref{fig:service}). Before a service client can invoke a service, it verifies the availability of a service server through a query to \codefont{wasm-js}, which tracks all available entities. The client publishes a request to the service request topic and awaits the server's response via the service response topic. The server publishes and subscribes to these topics in reverse. Figure \ref{fig:classes} illustrates the relationships between the participants as implemented in the \codefont{wasm-cpp} package.

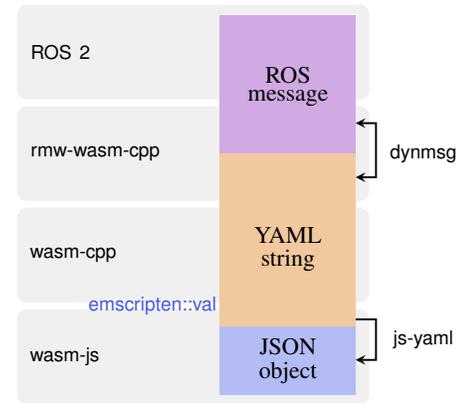
\begin{figure}[t]
    \centering
    \scalebox{0.9}{
      \begin{tikzpicture}
          \node (rosNode) [
              box,
              yshift = 2.25cm,
          ] {\codefont{ROS 2}};
          \node (rmw) [
              box,
              yshift = .75cm,
          ] {\codefont{rmw-wasm-cpp}};
          \node (cpp) [
              box,
              yshift = -.75cm,
          ] {\codefont{wasm-cpp}};
          \node (js) [
              box,
              yshift = -2.25cm,
          ] {\codefont{wasm-js}};
          \node (emval) [
              rectangle,
              yshift = -1.5cm,
              xshift = -0.6cm,
          ] {\color{colBlue}\codefont{emscripten::val}};

          \node (ros) [
              msg,
              minimum height = 2.1cm,
              yshift = 2.8cm,
              fill = colPink!80,
          ] {\shortstack{ROS \\ message}};
          \node (yaml) [
              msg,
              minimum height = 2.8cm,
              yshift = 0.75cm,
              fill = colOrange!50,
          ] {\shortstack{YAML \\ string}};
          \node (json) [
              msg,
              minimum height = 1cm,
              yshift = -1.8cm,
              fill = colBlue!40,
          ] {\shortstack{JSON \\ object}};

          \node (dyn) [
              rectangle,
              xshift = 3.4cm,
              yshift = 0.75cm,
              text centered,
          ] {\codefont{dynmsg}};

          \node (jsyaml) [
              rectangle,
              xshift = 3.4cm,
              yshift = -2cm,
          ] {\codefont{js-yaml}};

          \draw [arrow, <->] (2.4,1.2) -- +(0.3,0) -- +(0.3,-0.8) -- +(0,-0.8);
          \draw [arrow, ->] (2.4,-1.7) -- +(0.3,0) -- +(0.3,-0.6) -- +(0,-0.6);
      \end{tikzpicture}
      }
      \caption{Message conversion flow within the \texttt{rmw-wasm} middleware. A ROS message is initially handled by \texttt{rmw-wasm-cpp}, where it is converted to a YAML string using dynamic message introspection (\texttt{dynmsg}). The YAML string is then passed to \texttt{wasm-cpp}, which bridges the communication to \texttt{wasm-js} via \texttt{emscripten::val}. In \texttt{wasm-js}, the YAML string is further converted into a JSON object using the \texttt{js-yaml} library. This flow ensures that ROS messages can be efficiently processed and transmitted within the browser-based WebAssembly environment.}
      \vspace*{-0.2cm}
      \label{fig:msgFlow}
  \end{figure}

Messages are transmitted in YAML format from participants to \codefont{wasm-js}, which resides on the browser's main thread (Figure~\ref{fig:msgFlow}). \codefont{wasm-js} is responsible for tracking network participants and managing message traffic. To support multiple ROS nodes in the browser, \codefont{wasm-js} spawns each node as a separate web worker. Web workers enable parallel execution of scripts in the background, allowing concurrent operations without affecting the performance of the main thread~\cite{musch2019new}.

Once the web workers are operational, \codefont{wasm-js} stores all incoming data from publishers in message stacks. These message stacks act as temporary storage, holding messages until they are requested by subscribers. The stored data is then sent back to subscribers upon request, ensuring efficient communication between nodes.

\subsection{Cross-compilation}
\label{sec:crosscompile}
The cross-compilation process leverages the RoboStack ecosystem~\cite{fischer2021robostack}, extending its tools for WebAssembly. The \codefont{vinca} tool, which retrieves package information such as versions and dependencies from \codefont{rosdistro} and generates build recipes for the \codefont{boa} build tool, was adapted to support the WASM target platform. This modification was relatively straightforward due to \codefont{vinca}'s existing support for multiple platforms, requiring only minimal changes to the build script generation.

All ROS 2 packages must be recompiled specifically for WebAssembly to function in the ROS2WASM environment. For external libraries and dependencies, our approach builds upon the \url{https://emscripten-forge.org}, which is planned to be incorporated into the conda-forge ecosystem in the future. Emscripten-forge already provides many commonly used external libraries (including Python packages and numerical toolboxes) pre-compiled for WebAssembly.%

As part of the development efforts for ROS2WASM, several dependencies were added to emscripten-forge, which sometimes necessitated the addition of patches tailored to WASM, e.g.~introducing preprocessor guards for Emscripten. These patches are intended to guide future users on best practices and common pitfalls when targeting WebAssembly.%

The overall build process closely follows that of RoboStack, with continuous integration set up to build and upload packages to an anaconda.org channel, available at \url{https://github.com/RoboStack/ros-humble}.

\subsection{Web Integration}
\label{sec:webintegration}
After building individual packages, the next challenge is serving them on a website. Our approach enables two distinct types of web deployments: For applications compiled from C/C++ source code (as shown in Figure~\ref{fig:c_to_wasm}), we directly cross-compile the code to a WASM module and serve it as a standalone web application (Figure~\ref{fig:vernie}). %
For writing and executing ROS 2 applications directly in the browser, we utilize JupyterLite. By specifying the required packages in an \codefont{environment.yaml} file and running \codefont{jupyter lite build}, a complete website is generated, built on top of JupyterLite, with all specified packages installed. Users can then run commands such as \codefont{import rclpy} in a Python environment within the browser, using \codefont{rclpy} as they would in a typical ROS 2 setup (Figure~\ref{fig:jupyterlite}). For those preferring C++, the \codefont{rclcpp} package can be used with the \codefont{xeus-cpp} kernel for JupyterLite.

\subsection{Robotics Toolbox for Python and Swift}
\label{sec:swift}
This section briefly reviews how non-ROS robotics applications can be adapted for WASM. While the Robotics Toolbox for Python required no changes, adapting the graphics-based Swift Robotics Simulator~\cite{corke2021not} to the sandboxed web environment meant addressing limitations in threading, networking, and file system access. We cross-compiled Swift to WebAssembly using Emscripten and the PyJS framework, which required three major modifications.

\begin{figure}[t]
    \centering
    \browserTab{Swift}
    \includegraphics[width=0.74\columnwidth]{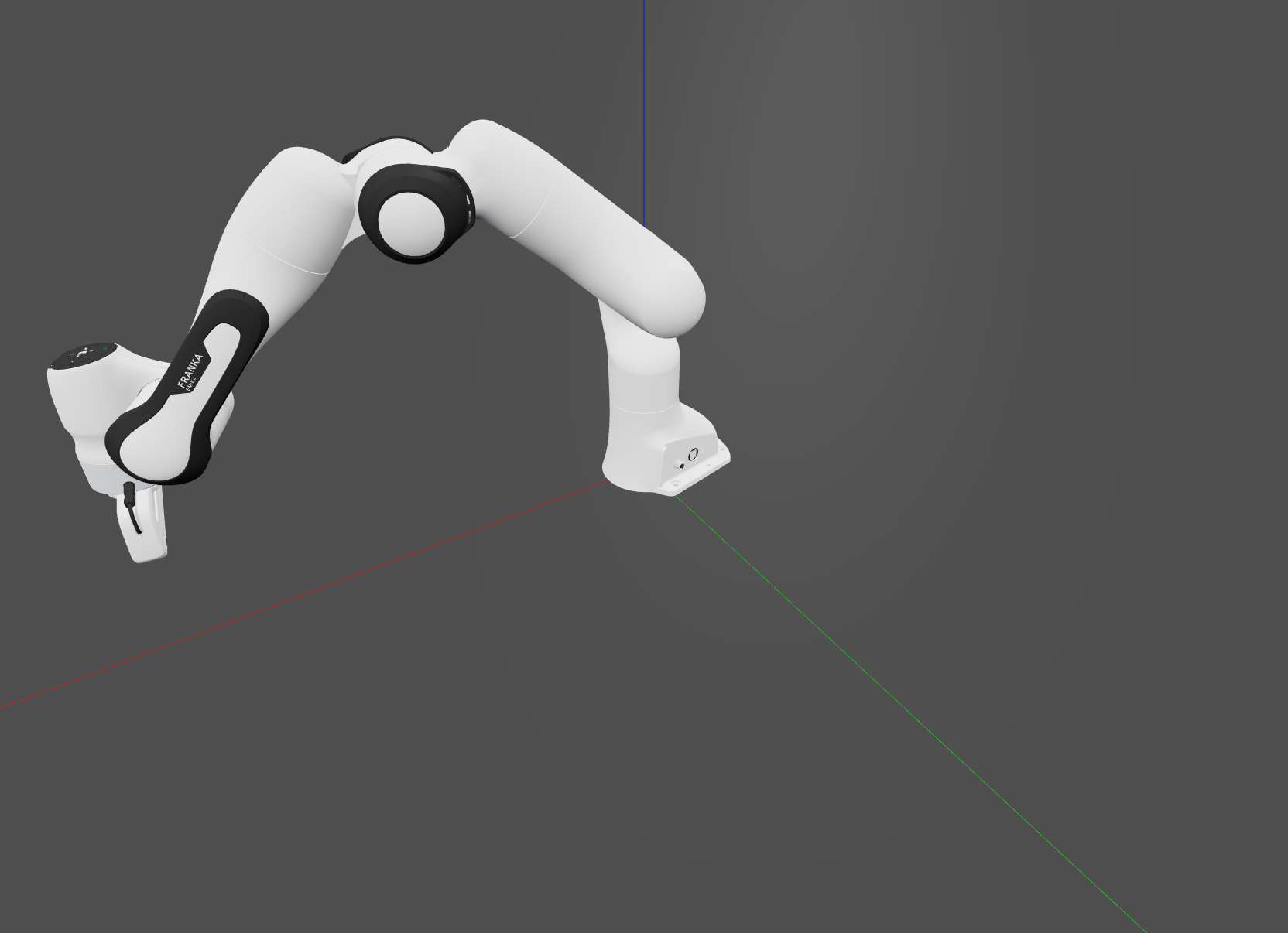}
    \caption{Swift is a lightweight browser-based simulator built on top of the Robotics Toolbox for Python. This simulator provides robotics-specific functionality for rapid prototyping of algorithms, research, and education. Here, we introduce Swift running in the browser without \textit{any} installation required: \url{www.ros2wasm.dev}.}
    \vspace*{-0.3cm}
    \label{fig:swift}
\end{figure}

First, we needed a new messaging system to replace WebSockets, which are not fully supported in WASM. This was achieved by passing messages via Python/Javascript interoperability using PyJS. Second, we had to find an alternative way to launch the Swift environment without starting an HTTP server. We eliminated the HTTP server by loading the web interface in an embedded iframe. Third, we needed to add hooks to load Robotics Toolbox assets from the packaged Python environment. This was achieved by intercepting asset fetching so that assets could be retrieved from the PyJS file system.

The resulting implementation demonstrates an approach for adapting existing projects to work within Emscripten and WASM constraints, which may inform future WebAssembly ports. Users can now interact with the simulator through a Python terminal embedded in a web page (Figure~\ref{fig:swift}).

\subsection{Communication with a non-ROS Robot}
\label{sec:vernie}
To showcase the wide range of possible applications of ROS2WASM, we demonstrate the manipulation of a physical robot. Data can be sent and received from a robot with the publishers and subscribers developed in Section~\ref{sec:middleware}, as long as there is a communication bridge between the robot and the web browser. There are two primary ways to communicate with a robot: Firstly, configure a robot to launch a rosbridge server to send information to the browser; however, this method requires the robot to be running ROS for creating a rosbridge. An alternative way for robots which do not run ROS natively is to create a bridge between the browser and the robot's system. The bridge can be established via Bluetooth, WiFi, or through a wired connection. Here, we demonstrate ROS2WASM with the LEGO BOOST Vernie robot that is connected to any computer via Bluetooth. A pre-existing library, \codefont{lego-boost-browser}~\cite{lego_boost_browser}, is used to establish the interface between ROS and Vernie via a Web Bluetooth API.

\begin{figure}[t]
    \centering
    \browserTab{LEGO Boost Vernie}
    \includegraphics[width=0.94\columnwidth]{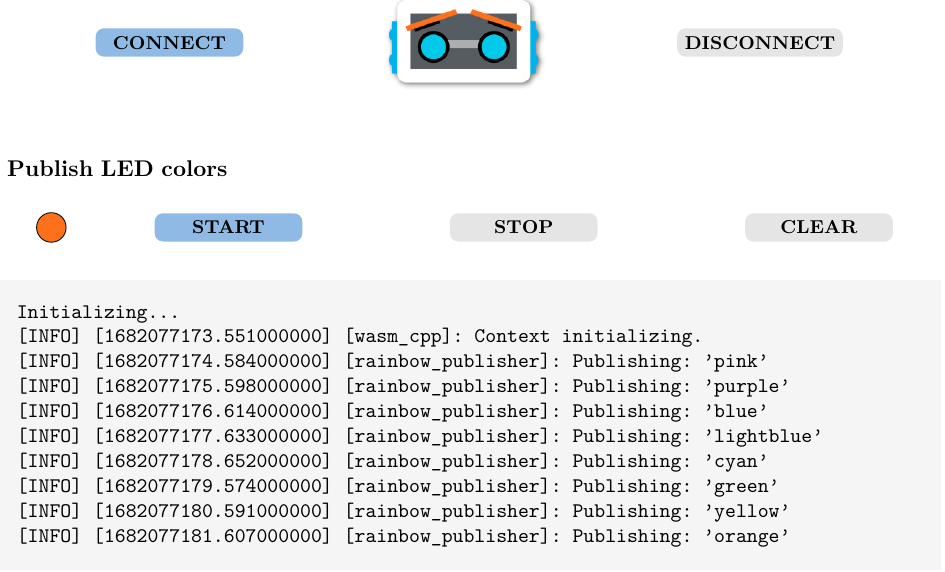}
    \vspace*{-.1cm}
    \caption{Interface for controlling the LEGO BOOST Vernie robot using a ROS publisher to send LED color commands. The user can connect or disconnect from the robot via Bluetooth and use the interface to publish different LED colors by starting or stopping the ``rainbow'' publisher. The console logs below display the sequence of colors being published in real time, illustrating the robot's response to the published messages.}
    \vspace*{-0.3cm}
    \label{fig:vernie}
\end{figure}

The \codefont{lego-boost-browser} library can send commands to Vernie to change the LED color or activate the two motors for locomotion. We created a publisher that simply publishes different LED colors as strings from Vernie's acceptable list of colors. When these messages reach the message stacks of the respective topic, they are forwarded to the robot through \codefont{lego-boost-browser}. Figure~\ref{fig:vernie} shows how a web page running an LED color publisher appears, displaying the user interface with connect/disconnect buttons, publisher controls, and a console that logs the sequence of colors being published in real-time to demonstrate the robot's response to commands. %
To control the position of the Vernie robot, we created a publisher that sends \codefont{geometry\_msgs/Twist} messages that contain linear and angular velocity vectors to express the robot's velocity in free space.

\section{Conclusion}
We demonstrated the feasibility and potential of running ROS 2 within a web browser using WebAssembly. Our work encompasses the design and implementation of three major components: a custom middleware architecture, a method for cross-compiling ROS 2 packages to WebAssembly, and the deployment of a web-based platform that allows users to interact with ROS environments directly in their browsers without the need for local installation or configuration.

Our website, \url{www.ros2wasm.dev}, enables the public to engage with a range of demonstrations that showcase the capabilities of ROS2WASM. These demonstrations vary in user interaction level and complexity, highlighting the versatility and scalability of the platform. By successfully running and facilitating communication between multiple ROS nodes within the browser, we have validated the core concept of ROS2WASM. This increased accessibility opens up new possibilities for education, research, and collaboration in robotics, making advanced tools more widely available. The current platform allows easy sharing of these environments via a simple link. However, further efforts are needed to simplify the setup process for novice users, particularly in creating and deploying their own ROS-based web applications.

ROS2WASM also has significant potential in industrial applications. Web-based ROS could enable robust server-based distributed control systems for multi-agent coordination, allowing operators to monitor and control robot fleets from any device without specialized software installations. Additionally, ROS2WASM could facilitate scalable and flexible cloud-based robotic processing, where computationally intensive tasks can be executed on remote servers while maintaining responsive browser-based interfaces on local devices.

Looking ahead, future work could explore integrating Zenoh as the middleware for ROS 2~\cite{zhang2023comparison}. Zenoh, a lightweight communication protocol recently added as an official RMW implementation for ROS 2, offers the potential to natively support WebAssembly. This could eliminate the need for our custom middleware, thereby streamlining the architecture and further improving performance and compatibility.

Another promising direction is adapting packages that contain graphical user interfaces such as the popular \codefont{rviz}, \codefont{Gazebo}, and \codefont{MoveIt} for browser compatibility. Such adaptations would further solidify the viability of running ROS 2 in the browser, which could lead to even broader adoption and innovation in web-based robotics.

\bibliographystyle{IEEEtran} 
\IEEEtriggeratref{14}
\bibliography{references}

\begin{thebibliography}{10}
\providecommand{\url}[1]{#1}
\csname url@samestyle\endcsname
\providecommand{\newblock}{\relax}
\providecommand{\bibinfo}[2]{#2}
\providecommand{\BIBentrySTDinterwordspacing}{\spaceskip=0pt\relax}
\providecommand{\BIBentryALTinterwordstretchfactor}{4}
\providecommand{\BIBentryALTinterwordspacing}{\spaceskip=\fontdimen2\font plus
\BIBentryALTinterwordstretchfactor\fontdimen3\font minus \fontdimen4\font\relax}
\providecommand{\BIBforeignlanguage}[2]{{%
\expandafter\ifx\csname l@#1\endcsname\relax
\typeout{** WARNING: IEEEtran.bst: No hyphenation pattern has been}%
\typeout{** loaded for the language `#1'. Using the pattern for}%
\typeout{** the default language instead.}%
\else
\language=\csname l@#1\endcsname
\fi
#2}}
\providecommand{\BIBdecl}{\relax}
\BIBdecl

\bibitem{macenski2022robot}
S.~Macenski, T.~Foote, B.~Gerkey, C.~Lalancette, and W.~Woodall, ``Robot operating system 2: Design, architecture, and uses in the wild,'' \emph{Science robotics}, vol.~7, no.~66, p. eabm6074, 2022.

\bibitem{estefo2019robot}
P.~Estefo, J.~Simmonds, R.~Robbes, and J.~Fabry, ``The robot operating system: Package reuse and community dynamics,'' \emph{Journal of Systems and Software}, vol. 151, pp. 226--242, 2019.

\bibitem{fischer2021robostack}
T.~Fischer, W.~Vollprecht, S.~Traversaro, S.~Yen, C.~Herrero, and M.~Milford, ``A {RoboStack} tutorial: Using the robot operating system alongside the conda and jupyter data science ecosystems,'' \emph{IEEE Robotics \& Automation Magazine}, vol.~29, no.~2, pp. 65--74, 2021.

\bibitem{cervera2018try}
E.~Cervera, ``Try to start it! the challenge of reusing code in robotics research,'' \emph{IEEE Robotics and Automation Letters}, vol.~4, no.~1, pp. 49--56, 2018.

\bibitem{kohut2023teaching}
M.~Koh{\'u}t, M.~{\v{C}}orn{\'a}k, M.~Dobi{\v{s}}, and A.~Babinec, ``Teaching robotics with the usage of {Robot Operating System ROS},'' in \emph{International Conference on Robotics in Education}, 2023, pp. 299--313.

\bibitem{cervera2023run}
E.~Cervera, ``Run to the source: The effective reproducibility of robotics code repositories,'' \emph{IEEE Robotics \& Automation Magazine}, 2023.

\bibitem{cervera2019roslab}
E.~Cervera and A.~P. Del~Pobil, ``{ROSLab: Sharing ROS Code Interactively With Docker and JupyterLab},'' \emph{IEEE Robotics \& Automation Magazine}, vol.~26, no.~3, pp. 64--69, 2019.

\bibitem{white2017ros}
R.~White and H.~Christensen, ``{ROS and Docker},'' \emph{Robot Operating System (ROS) The Complete Reference (Volume 2)}, pp. 285--307, 2017.

\bibitem{haas2017bringing}
A.~Haas, A.~Rossberg, D.~L. Schuff, B.~L. Titzer, M.~Holman, D.~Gohman, L.~Wagner, A.~Zakai, and J.~Bastien, ``Bringing the web up to speed with {WebAssembly},'' in \emph{ACM SIGPLAN Conference on Programming Language Design and Implementation}, 2017, pp. 185--200.

\bibitem{wang2021empowering}
W.~Wang, ``Empowering web applications with {WebAssembly}: Are we there yet?'' in \emph{IEEE/ACM International Conference on Automated Software Engineering}, 2021, pp. 1301--1305.

\bibitem{quigley2009ros}
M.~Quigley, K.~Conley, B.~Gerkey, J.~Faust, T.~Foote, J.~Leibs, R.~Wheeler, A.~Y. Ng \emph{et~al.}, ``{ROS: an open-source Robot Operating System},'' in \emph{IEEE International Conference on Robotics and Automation Workshop on Open Source Software}, 2009.

\bibitem{hellmund2016robot}
A.-M. Hellmund, S.~Wirges, {\"O}.~{\c{S}}. Ta{\c{s}}, C.~Bandera, and N.~O. Salscheider, ``Robot operating system: A modular software framework for automated driving,'' in \emph{IEEE International Conference on Intelligent Transportation Systems}, 2016, pp. 1564--1570.

\bibitem{he2022ros2}
J.~He, J.~Zhang, J.~Liu, and X.~Fu, ``A {ROS2-based} framework for industrial automation systems,'' in \emph{International Conference on Computer, Control and Robotics}, 2022, pp. 98--102.

\bibitem{angrybots}
J.~Echterhoff, ``Angry bots demo,'' \url{https://beta.unity3d.com/jonas/AngryBots/}, accessed: 2024-09-10.

\bibitem{d3wasm}
{Continuation Labs}, ``D3wasm project,'' \url{https://www.continuation-labs.com/projects/d3wasm/}, accessed: 2024-09-10.

\bibitem{zakai2011emscripten}
A.~Zakai, ``{Emscripten: an LLVM-to-JavaScript compiler},'' in \emph{ACM International Conference Companion on Object Oriented Programming Systems Languages and Applications Companion}, 2011, pp. 301--312.

\bibitem{LLVM}
C.~Lattner and V.~Adve, ``{LLVM}: A compilation framework for lifelong program analysis and transformation,'' in \emph{International Symposium on Code Generation and Optimization}, 2004, p. 75–88.

\bibitem{rosonweb}
M.~Allwright, ``{ROS On Web},'' \url{https://rosonweb.io/}, accessed: 2024-09-10.

\bibitem{roswasm_suite}
N.~Bore, ``roswasm\_suite: Libraries for compiling {C++ ROS} nodes to {Webassembly} using {Emscripten},'' \url{https://github.com/nilsbore/roswasm_suite}, accessed: 2024-09-10.

\bibitem{rosweb}
M.~A.~N. da~Cunha~de Arruda, ``{ROSWeb: A} web based supervisory system for {ROS},'' \url{https://github.com/EESC-LabRoM/rosweb}, accessed: 2024-09-10.

\bibitem{foxglove}
{Foxglove Technologies Inc.}, ``Foxglove studio: Robotics visualization and debugging tool,'' \url{https://foxglove.dev/}, accessed: 2024-09-10.

\bibitem{crick2017rosbridge}
C.~Crick, G.~Jay, S.~Osentoski, B.~Pitzer, and O.~C. Jenkins, ``{Rosbridge: ROS for Non-ROS Users},'' in \emph{International Symposium on Robotics Research}, 2017, pp. 493--504.

\bibitem{ros_control_center}
L.~Berscheid, ``Ros control center: A universal tool for controlling robots running {ROS},'' \url{https://github.com/pantor/ros-control-center}, accessed: 2024-09-10.

\bibitem{rosboard}
D.~Venkatraman, ``Rosboard: A {ROS} node that runs a web server on your robot,'' \url{https://github.com/dheera/rosboard}, accessed: 2024-09-10.

\bibitem{ros_over_email}
C.~Bédard, ``{ROS 2 over Email: rmw\_email, an actual working RMW implementation},'' \url{https://christophebedard.com/ros-2-over-email/}, accessed: 2024-09-10.

\bibitem{musch2019new}
M.~Musch, C.~Wressnegger, M.~Johns, and K.~Rieck, ``New kid on the web: A study on the prevalence of webassembly in the wild,'' in \emph{International Conference on the Detection of Intrusions and Malware, and Vulnerability Assessment}, 2019, pp. 23--42.

\bibitem{corke2021not}
P.~Corke and J.~Haviland, ``{Not your grandmother’s toolbox--the Robotics Toolbox reinvented for Python},'' in \emph{IEEE International Conference on Robotics and Automation}, 2021, pp. 11\,357--11\,363.

\bibitem{lego_boost_browser}
T.~Tuhkanen, ``{Lego Boost Browser: Control Lego Boost from the browser},'' \url{https://github.com/ttu/lego-boost-browser}, accessed: 2024-09-10.

\bibitem{zhang2023comparison}
J.~Zhang, X.~Yu, S.~Ha, J.~P. Queralta, and T.~Westerlund, ``{Comparison of DDS, MQTT, and Zenoh in Edge-to-Edge/Cloud Communication with ROS 2},'' \emph{arXiv preprint arXiv:2309.07496}, 2023.

\end{thebibliography}

\end{document}